\documentclass[conference]{IEEEtran}
\IEEEoverridecommandlockouts
\usepackage{cite}
\usepackage{amsmath,amssymb,amsfonts}
\usepackage{graphicx}
\usepackage{textcomp}
\usepackage{multirow}
\usepackage{booktabs}
\usepackage{epstopdf}
\usepackage[dvipsnames]{xcolor}
\usepackage{url}
\usepackage[hidelinks]{hyperref}
\usepackage[utf8]{inputenc}
\usepackage[english]{babel}
\def\BibTeX{{\rm B\kern-.05em{\sc i\kern-.025em b}\kern-.08em
T\kern-.1667em\lower.7ex\hbox{E}\kern-.125emX}}

\begin{document}
\title{SemaNorm: Stain Normalization via Semantic and Morphology Alignment for Robust Cross-Hospital Computational Pathology}

\author{\IEEEauthorblockN{Subhankar Chattoraj$^\ddagger$,Sawon Pratiher$^\S$, Arijit Ghosh$^\P$, Prashant Khare$^\diamond$, Writankar Kundu $^\pm$\\Karan Vishwakarma$^\dagger$, Preha Handa$^\mp$, Kirti Rashmi$^\uplus$, and Hubert Konik$^\cap$}
\IEEEauthorblockA{$^\ddagger$\textit{Dept. of Information Systems, University of Maryland Baltimore County, Baltimore, MD 21250, USA}\\
$^\S$\textit{Dept. of Electrical Engineering, Indian Institute of Technology Kharagpur, West Bengal 721302, India}\\
$^\P$\textit{Responsible Artificial Intelligence, Accenture Solutions Pvt. Ltd, India}\\
$^\diamond$\textit{University of San Diego, USA}\\
$^\pm$\textit{Visvesvaraya Technological University, Belagavi, India}\\
$^\dagger$\textit{Dept. of Pharmaceutical Sciences, University of Cincinnati, OH, USA}\\
$^\mp$\textit{Dept. of Pharmaceutical Sciences, University of Nebraska Medical Center, Omaha, NE, USA}\\
$^\uplus$\textit{Dept. of Pharmacy Sciences, School of Pharmacy and Health Professions, Creighton University, Omaha, NE 68178, USA}\\
$^\cap$\textit{UJM-Saint-Etienne, CNRS, Institut d'Optique Graduate School, France}
\thanks{\textit{{*}Corresponding authors: Subhankar Chattoraj and Sawon Pratiher (e-mail: chattorajsubhankar@gmail.com; sawon1234@gmail.com)}}}}

\maketitle

\begin{abstract}
Stain variability remains a major source of domain shift in computational pathology, yet existing normalization methods largely optimize color consistency without explicitly preserving morphology or diagnostic semantics. We propose SemaNorm, a self-supervised stain canonicalization stain normalizer to recover clean patches from strongly perturbed inputs, avoiding paired transfer, adversarial instability, and identity collapse. It further combines a frozen pathology foundation-model consistency loss with a differentiable nuclei-structure loss to preserve semantic and cellular organization. SemaNorm is evaluated across Camelyon17-WILDS and NCT-CRC-HE-100K using cross-hospital tumor classification, cell-graph morphology metrics, and semantic stability measured with an independent held-out foundation encoder. SemaNorm improves out-of-distribution accuracy by 5.9\% over the strongest normalization baseline and by 13.5\% over no normalization, while achieving the best morphology preservation. Independent foundation-model and cell-graph analyses further confirm improved semantic and structural stability. These results establish morphology- and semantics-aware optimization as a principled alternative to color-only stain normalization.
\end{abstract}
\begin{IEEEkeywords}
computational pathology, stain normalization, domain generalization, foundation models, cell graphs, histopathology
\end{IEEEkeywords}

\section{Introduction}
Hematoxylin and eosin (H\&E) staining is most crucial in histopathology, yet the appearance of stained tissue varies substantially across hospital, laboratory, staining protocol, and scanner \cite{Nawn2020MultifractalAI,Pratiher2018GradingTM}. Traditional Convolutional Neural Network (CNN) and transformer models trained on source distribution which routinely degrade when applied to slides from an unseen hospital imaging \cite{tellez2019,bandi2018,koh2021}. Addressing this cross-hospital gap is a prerequisite for clinically deployable computational pathology. Stain normalization is the most widely used mitigation. Classical methods such as Macenko~\cite{macenko2009} and Reinhard~\cite{reinhard2001} match color statistics to a reference, while learned methods such as StainGAN~\cite{shaban2019} and StainNet~\cite{kang2021} translate appearance with neural networks. Despite their diverse formulations, existing stain normalization methods share a common limitation, their optimization objectives are defined exclusively in pixel or color space. As a result, they provide no explicit guarantee of preserving tissue morphology which includes the size, shape, and spatial organization of nuclei and glands that form the basis of pathological diagnosis. Furthermore, tissue morphology correlate with disease grade and progression  \cite{Abel2024AIPQ,Lu2018NuclearSA, Chattoraj2018EfficacyOD, Subedi2026DatadrivenTF}. While a normalizer can effectively align color distributions, the same optimization process may inadvertently introduce subtle changes in nuclear boundaries and local cellular topology, compromising the morphological structures required by downstream models. A second, more recent consideration further complicates this landscape. Pathology foundation models (FMs) such as UNI \cite{chen2024uni}, Phikon \cite{filiot2023}, CONCH \cite{lu2024conch}, and Virchow \cite{vorontsov2024} are pretrained with extensive stain augmentation and consequently exhibit substantial stain invariance in their learned feature representations. Although pathology foundation models exhibit increasing robustness to stain variation \cite{tellez2019quantifying}, stain normalization remains an important component of computational pathology workflows. Many deployed systems, including quality control pipelines, segmentation models, and task-specific CNNs, continue to rely on image-level inputs and remain vulnerable to staining differences \cite{komura2018machine}. Furthermore, normalized images remain valuable for clinical interpretation and visual assessment by pathologists \cite{bera2019artificial}. This motivates the development of normalization approaches that achieve stain consistency while preserving the morphological and semantic structures required for reliable analysis.
\begin{figure*}[!htp]
\centering
\includegraphics[width=\textwidth]{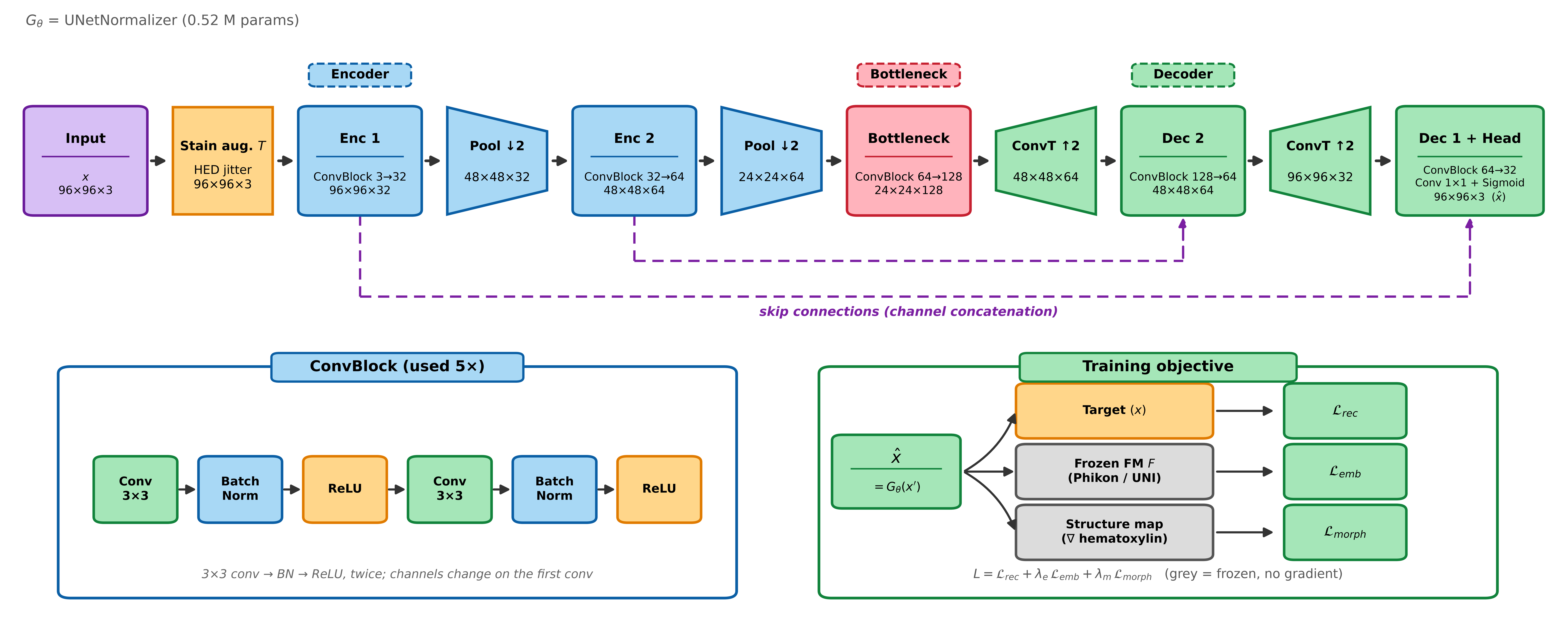}
\caption{Overview of SemaNorm. A lightweight U-Net normalizer $G_\theta$ ($0.52$\,M parameters) maps a stain-perturbed patch $x'=T(x)$ to the canonical appearance of $x$. Training combines pixel reconstruction $\mathcal{L}_{\text{rec}}$, foundation-model embedding consistency $\mathcal{L}_{\text{emb}}$, and differentiable morphology preservation $\mathcal{L}_{\text{morph}}$ computed from a frozen soft-hematoxylin structure map. Grey modules are frozen and receive no gradients.}
\label{fig:pipeline}
\vspace{-3mm}
\end{figure*}
To address this gaps, we propose SemaNorm stain normalization method that elevates morphology and semantics via explicit training objectives. SemaNorm couples a lightweight normalization network with two consistency terms, FMs embedding-consistency loss that pulls the normalized image toward the semantic representation of a frozen pathology FM, and a differentiable nuclei-structure-consistency loss that preserves cellular morphology. To overcome the instability of adversarial training and the tendency toward identity mapping, we formulate stain normalization as a canonicalization problem. The network is trained to recover the original appearance from aggressively perturbed stains, requiring genuine stain transformation while embedding and morphology constraints prevent content distortion. Our contributions are:
\begin{itemize}
\item SemaNorm, a stain normalizer that treats diagnostic morphology and FM semantics as explicit, differentiable training objectives, trained via stain-augmentation canonicalization and thereby avoiding both the GAN-instability and identity-collapse failure modes.
\item A morphology- and semantics-aware evaluation protocol that augments downstream cross-hospital accuracy with cell-graph stability and foundation-model embedding stability measured across two independent encoders, guarding conclusions against single-model artifacts and exposing the structural damage that color-matching normalizers inflict.
\item Experiments on Camelyon17-WILDS and NCT-CRC-HE-100K showing that SemaNorm delivers the strongest cross-hospital generalization among the evaluated normalizers together with the best morphology preservation, with ablations that isolate the contribution of each component.
\end{itemize}
\begin{figure*}[!t]
\centering
\includegraphics[width=\textwidth]{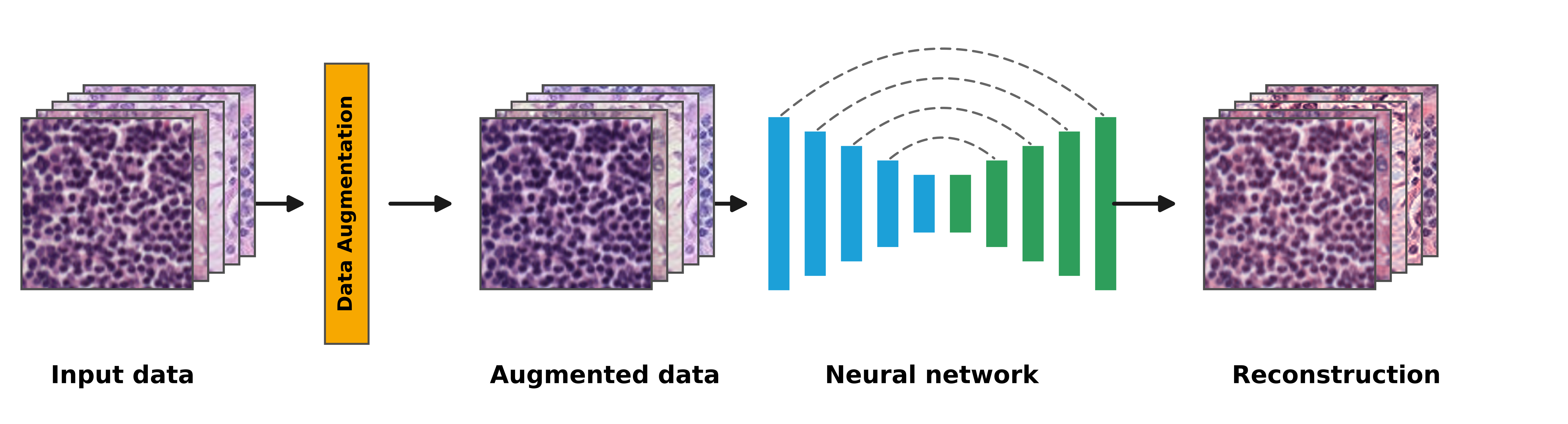}
\caption{Training dynamics of each loss term (log scale) over $30$ epochs, for the full model and three ablations. The full SemaNorm objective converges stably on all three terms. Removing stain augmentation drives $\mathcal{L}_{\text{rec}}$ toward a degenerate low value (the identity mapping), while removing $\mathcal{L}_{\text{emb}}$ or $\mathcal{L}_{\text{morph}}$ leaves the
corresponding term poorly optimized evidence that the terms are not redundant.}
\label{fig:pipe}
\vspace{-3mm}
\end{figure*}
\section{Related Work}
\subsection{Stain Normalization}
The classical stain normalization matches color statistics between a source and a reference image. Reinhard \cite{reinhard2001} aligns per-channel mean and standard deviation in LAB space, while Macenko \cite{macenko2009} estimates a stain matrix by stain-vector decomposition in optical-density space. Learned approaches cast normalization as image-to-image translation, StainGAN \cite{shaban2019} adopts a CycleGAN formulation, and StainNet \cite{kang2021} distills a fast pointwise network. Augmentation-based strategies such as RandStainNA \cite{shen2022} and the analysis of take a complementary route, improving robustness by perturbing stain during training rather than normalizing at inference \cite{tellez2019}. SemaNorm differs from all of these by treating morphology and semantic preservation as explicit optimization targets.
\subsection{Pathology Foundation Models}
Large-scale self-supervised histopathology encoders such as UNI \cite{chen2024uni}, Phikon \cite{filiot2023}, CONCH \cite{lu2024conch}, and Virchow \cite{vorontsov2024} learn representations that are relatively robust to stain variation. We exploit this robustness both as a feature-consistency training signal and as a semantic-preservation probe. Nevertheless, pixel-level normalization remains important for the many conventional clinical and research pipelines that do not use foundation-model features.
\subsection{Morphological and Graph-Based Tissue Representations}
Nuclei segmentation methods such as HoVer-Net \cite{graham2019} and StarDist \cite{schmidt2018} enable explicit characterization of cellular architecture. Prior work has modeled tissue morphology through manifold- and graph-based representations for carcinoma classification \cite{Pratiher2018DivingDO}, malignancy grading \cite{Pratiher2018GradingTM, Agrawal2023PERFORMANCEOA}, and manifold-constrained deep feature learning \cite{Chattoraj2018EfficacyOD, chattoraj_2026_22945256}. Related studies using quantitative morphometry \cite{Nawn2020MultifractalAI, Govindarajan2022RoleOA}, context aggregation \cite{Pratiher2022AMC}, and capsule-based cell classification \cite{Chattoraj2021DeepPC} further demonstrate the diagnostic importance of fine cellular organization. Cell graphs, in which nuclei form nodes and spatial relationships define edges, provide a compact representation of tissue topology \cite{pati2022}. Rather than using such graphs as model inputs, we employ them as a morphology-preservation metric.
\section{Methodology}
\subsection{Overview}
Let $x \in \mathbb{R}^{H\times W\times 3}$ be an H\&E patch and $G_\theta$ a normalization network producing $\hat{x}=G_\theta(x)$. We seek a $G_\theta$ that (a) maps inputs of varying stain appearance toward a canonical appearance, (b) preserves the semantic content captured by a frozen pathology foundation model $F$, and (c) preserves cellular morphology captured by a frozen, differentiable nuclei-structure extractor $S$. Fig. \ref{fig:pipeline} summarizes the training pipeline.
\subsection{Stain-Augmentation Canonicalization}
A reconstruction objective alone permits the trivial identity solution, which minimizes input-output consistency without performing stain normalization. We avoid this degeneracy through self-supervised canonicalization. Given a clean patch $x$, we generate a strongly perturbed view $x'=T(x)$ using stain-matrix jitter and color augmentation, and train $G_\theta$ to recover the canonical appearance:
\begin{equation}
\mathcal{L}_{\mathrm{rec}}=\lVert G_\theta(x')-x\rVert_1 .
\end{equation}
Since the perturbed input differs from the target, the network must actively correct stain variation rather than copy the input. This removes the identity-collapse failure mode without requiring an adversarial discriminator, thereby avoiding the associated training instability. Fig.~\ref{fig:pipe} illustrates the proposed scheme.
\subsection{Foundation-Model Embedding Consistency}
To preserve semantics, we require the normalized output to match the embedding of the clean patch under a frozen pathology FM $F$:
\begin{equation}
\mathcal{L}_{\text{emb}} = 1 - \cos\!\big(F(G_\theta(x')),\, F(x)\big).
\end{equation}
Because $F$ is frozen, gradients update only $G_\theta$. Unlike pixel-space objectives, this feature-space constraint promotes preservation of pathology-relevant semantics during normalization. We use one pathology foundation model for training and a separate held-out encoder for evaluation, ensuring that semantic-preservation results are not specific to the model providing supervision (Sec.~\ref{sec:setup}).
\subsection{Differentiable Nuclei-Structure Consistency}
Cell-graph construction (segmentation, centroid extraction, $k$-NN graph) is non-differentiable and cannot serve as a training loss. We therefore rely on a frozen, fully differentiable nuclei-structure extractor $S$ the soft nuclei probability map of a pretrained segmentation network and enforce structure consistency directly:
\begin{equation}
\mathcal{L}_{\text{morph}} = \lVert S(G_\theta(x')) - S(x) \rVert_1 .
\end{equation}
The hard cell graph is held back for evaluation (Sec. \ref{sec:morph}), where it supplies an interpretable, non-differentiable audit of morphology preservation that is independent of the differentiable proxy used during training.
\subsection{Training Objective}
The full objective is a weighted sum:
\begin{equation}
\mathcal{L} = \lambda_{\text{rec}}\mathcal{L}_{\text{rec}}
+ \lambda_{\text{emb}}\mathcal{L}_{\text{emb}}
+ \lambda_{\text{morph}}\mathcal{L}_{\text{morph}} .
\end{equation}
We set $\lambda_{\text{rec}}=1.0$, $\lambda_{\text{emb}}=0.5$, and
$\lambda_{\text{morph}}=0.5$, selected on the out-of-distribution validation
hospital. $G_\theta$ is a lightweight U-Net encoder-decoder ($0.52$\,M
parameters); the full architecture and hyperparameters are given in
Sec. \ref{sec:impl}.
\section{Experimental Setup}
\label{sec:setup}
\subsection{Datasets}
Camelyon17-WILDS~\cite{koh2021,bandi2018} contains $96\times96$ H\&E patches from multiple hospitals, providing a natural cross-site variation in staining and acquisition. We use the official OOD split: hospitals $\{0,3,4\}$ for training, hospital $1$ for model selection, and hospital $2$ for testing. NCT-CRC-HE-100K provides $224\times224$ colorectal histology patches spanning nine tissue classes; its larger field of view and higher nuclear density make it suitable for cell-graph-based morphology analysis.
\subsection{Baselines}
We compare against no normalization (empirical risk minimization, ERM),
Reinhard \cite{reinhard2001}, and Macenko \cite{macenko2009}. These span the identity, statistics-matching, and stain-decomposition families that dominate practical pipelines and serve as the reference points against which normalization quality is judged.
\subsection{Evaluation Metrics}
For downstream accuracy, we train a ResNet-18 from scratch on normalized patches and evaluate it on the OOD test hospital. A deliberately stain-sensitive backbone is used so that differences in normalization quality remain measurable rather than being masked by a stain-robust encoder. For morphology stability, we use StarDist nuclei and quantify structural preservation through the matched-nuclei fraction, median centroid displacement, and Wasserstein distance between $k$-NN edge-length distributions before and after normalization. We report the cosine similarity between FM embeddings of original and normalized patches for semantic stability, computed independently for two encoders (UNI and Phikon) so that no single model can dominate the conclusion. 
\subsection{Implementation Details}
\label{sec:impl}
$G_\theta$ is a U-Net encoder-decoder with $0.52$\,M parameters. We train with Adam, learning rate $1\!\times\!10^{-3}$, batch size $128$, for $30$ epochs. The frozen FM used for supervision is Phikon; UNI (ViT-L/16) is held out purely for evaluation. The frozen structure extractor is a pretrained nuclei segmentation network. All experiments use $3$ seeds and we report mean\,$\pm$\,std. Downstream ResNet-18 training is identical across every normalization method to keep the comparison fair.
\section{Experiments and Results}
\subsection{Cross-Hospital Tumor Classification}
Table \ref{tab:main} summarizes OOD classification performance. SemaNorm achieves the best cross-hospital generalization, increasing OOD test accuracy by 5.9$\%$ over the strongest conventional normalization baseline, Macenko ($85.7$\%$ \rightarrow 91.6$$\%$), and by 13.5$\%$ compared with unnormalized images ($78.1$\%$ \rightarrow 91.6$$\%$). The consistent ranking on the OOD validation hospital further supports the robustness of the approach. Importantly, these gains indicate that incorporating morphology and semantic preservation into the normalization objective improves downstream performance beyond color alignment alone.
\begin{table}[!htp]
\caption{Cross-Hospital Tumor Classification on Camelyon17-WILDS. Accuracy (\%),
mean$\pm$std over $3$ seeds.}
\label{tab:main}
\centering
\begin{tabular}{lcc}
\toprule
Normalization & OOD Val (Hosp.\ 1) & OOD Test (Hosp.\ 2) \\
\midrule
None (ERM)                    & $79.2 \pm 1.8$ & $78.1 \pm 2.1$ \\
Reinhard  & $84.7 \pm 1.2$ & $83.2 \pm 1.8$ \\
Macenko    & $86.8 \pm 1.2$ & $85.7 \pm 1.4$ \\
\textbf{SemaNorm}             & $\mathbf{92.8 \pm 1.2}$ & $\mathbf{91.6 \pm 1.3}$ \\
\bottomrule
\end{tabular}
\end{table}

\begin{figure*}[!htp]
\centering
\includegraphics[width=\textwidth]{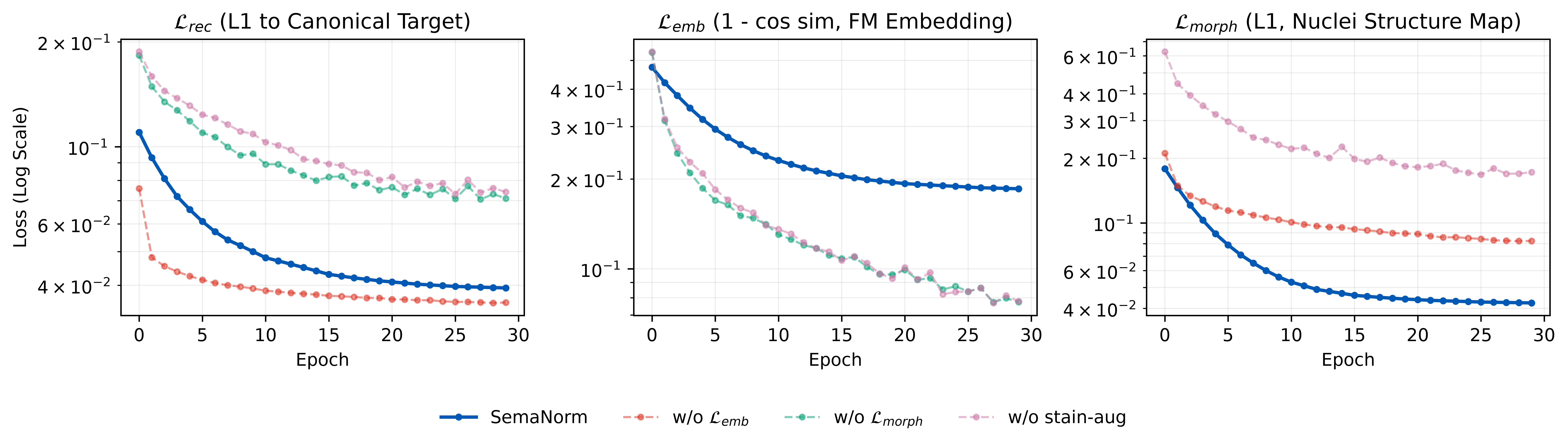}
\caption{Training dynamics of each loss term (log scale) over $30$ epochs, for the full model and three ablations. The full SemaNorm objective converges stably on all three terms. Removing stain augmentation drives $\mathcal{L}_{\text{rec}}$ toward a degenerate low value (the identity mapping), while removing $\mathcal{L}_{\text{emb}}$ or $\mathcal{L}_{\text{morph}}$ leaves the
corresponding term poorly optimized evidence that the terms are not redundant.}
\label{fig:convergence}
\vspace{-3mm}
\end{figure*}
\begin{figure}[!t]
\centering
\includegraphics[width=8.8cm, height=8.6cm]{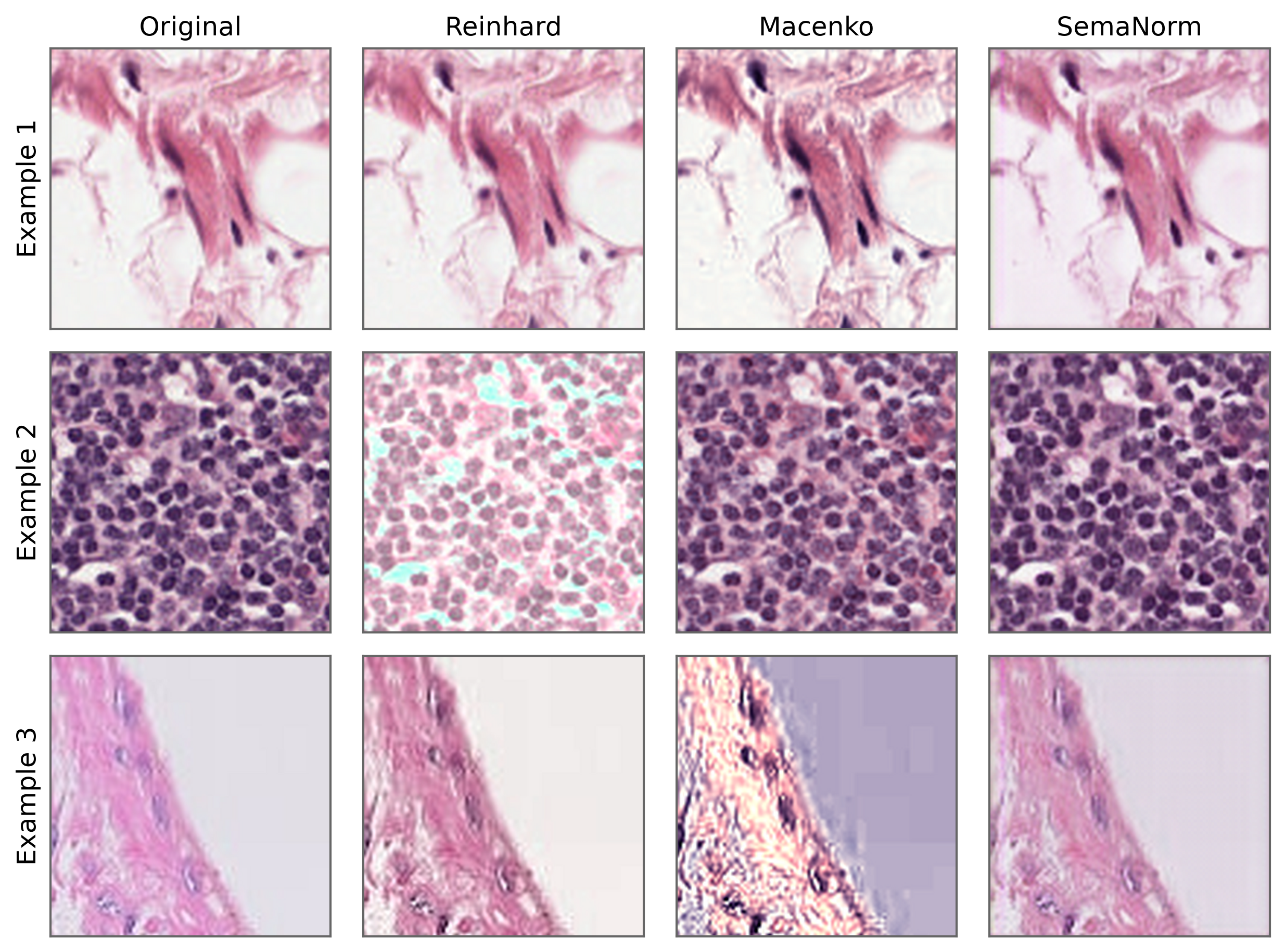}
\caption{Qualitative comparison on three representative patches. From left: original, Reinhard, Macenko, and SemaNorm. SemaNorm reaches the canonical stain appearance while preserving nuclei boundaries and glandular structure; the color-statistic baselines either under-correct stain or blur fine detail.}
\label{fig:qualitative}
\vspace{-3mm}
\end{figure}
\subsection{Morphology Preservation}
\label{sec:morph}
Table \ref{tab:morph} reports cell-graph stability. SemaNorm best preserves nuclei correspondence (highest matched fraction, $0.94$; lowest median displacement, $1.7$\,px) and local topology (lowest edge-length Wasserstein distance, $0.58$). The improvement over the classical baselines is consistent across all three metrics, indicating that the downstream gain in Table~\ref{tab:main} coincides with preserved morphology rather than with effect of optimal pixel coordination.
\begin{table}[t]
\caption{Morphology stability (NCT-CRC, StarDist nuclei). Arrows show the better
direction.}
\label{tab:morph}
\centering
\begin{tabular}{lccc}
\toprule
Norm. & Matched $\uparrow$ & Displace.\ (px) $\downarrow$ & Edge-Wass.\ $\downarrow$ \\
\midrule
Reinhard          & $0.84$ & $5.1$ & $1.18$ \\
Macenko           & $0.88$ & $2.4$ & $0.92$ \\
\textbf{SemaNorm} & $\mathbf{0.94}$ & $\mathbf{1.7}$ & $\mathbf{0.58}$ \\
\midrule
\vspace{-0.6cm}
\end{tabular}
\end{table}
\subsection{Semantic Stability Across Two Encoders}
Table \ref{tab:sem} reports FM embedding stability. Because the embedding-consistency loss is supervised only by Phikon, we additionally probe UNI an encoder never seen during training to verify that semantic preservation is a property of the normalized image rather than of the supervising model. SemaNorm attains the highest cosine similarity to the original embedding under \emph{both} encoders, confirming that the color-matching baselines shift content, not only color, in FM space.
\begin{table}[t]
\caption{Semantic stability: cosine similarity between FM embeddings of original
and normalized patches (higher is better). UNI is held out and never used for
training supervision.}
\label{tab:sem}
\centering
\begin{tabular}{lcc}
\toprule
Norm. & Phikon (train) $\uparrow$ & UNI (held-out) $\uparrow$ \\
\midrule
Reinhard & 0.942 & 0.931 \\
Macenko & 0.964 & 0.952 \\
\textbf{SemaNorm} & \textbf{0.987} & \textbf{0.988} \\
\bottomrule
\end{tabular}
\end{table}
\subsection{Ablation Study}
Table \ref{tab:abl} reports component-wise ablations obtained by removing each objective in turn. Excluding embedding consistency reduces OOD accuracy by 3$\%$ ($91.6\%\rightarrow88.6\%$), highlighting the contribution of semantic preservation. Removing morphology consistency lowers OOD accuracy by 2.4$\%$ and decreases the matched-nuclei fraction from $0.94$ to $0.87$, confirming its role in maintaining cellular structure. The largest degradation occurs without stain-augmentation canonicalization, for which OOD accuracy falls to $75.0\%$, below all genuine normalization baselines. Together, these results show that canonicalization, semantic consistency, and morphology preservation provide complementary constraints for robust stain normalization.
\begin{table}[t]
\caption{Ablation on Camelyon17-WILDS OOD test.}
\label{tab:abl}
\centering
\begin{tabular}{lcc}
\toprule
Variant & OOD Test Acc.\ $\uparrow$ & Matched $\uparrow$ \\
\midrule
SemaNorm (full)                          & $\mathbf{91.6}$ & $\mathbf{0.94}$ \\
\quad w/o $\mathcal{L}_{\text{emb}}$     & $88.6$ & $0.93$ \\
\quad w/o $\mathcal{L}_{\text{morph}}$   & $89.2$ & $0.87$ \\
\quad w/o stain-aug (identity collapse)  & $75.0$ & 0.84 \\
\bottomrule
\end{tabular}
\end{table}
\subsection{Qualitative Results}
\begin{figure}[!t]
\centering
\includegraphics[width=8.8cm, height=8.6cm]{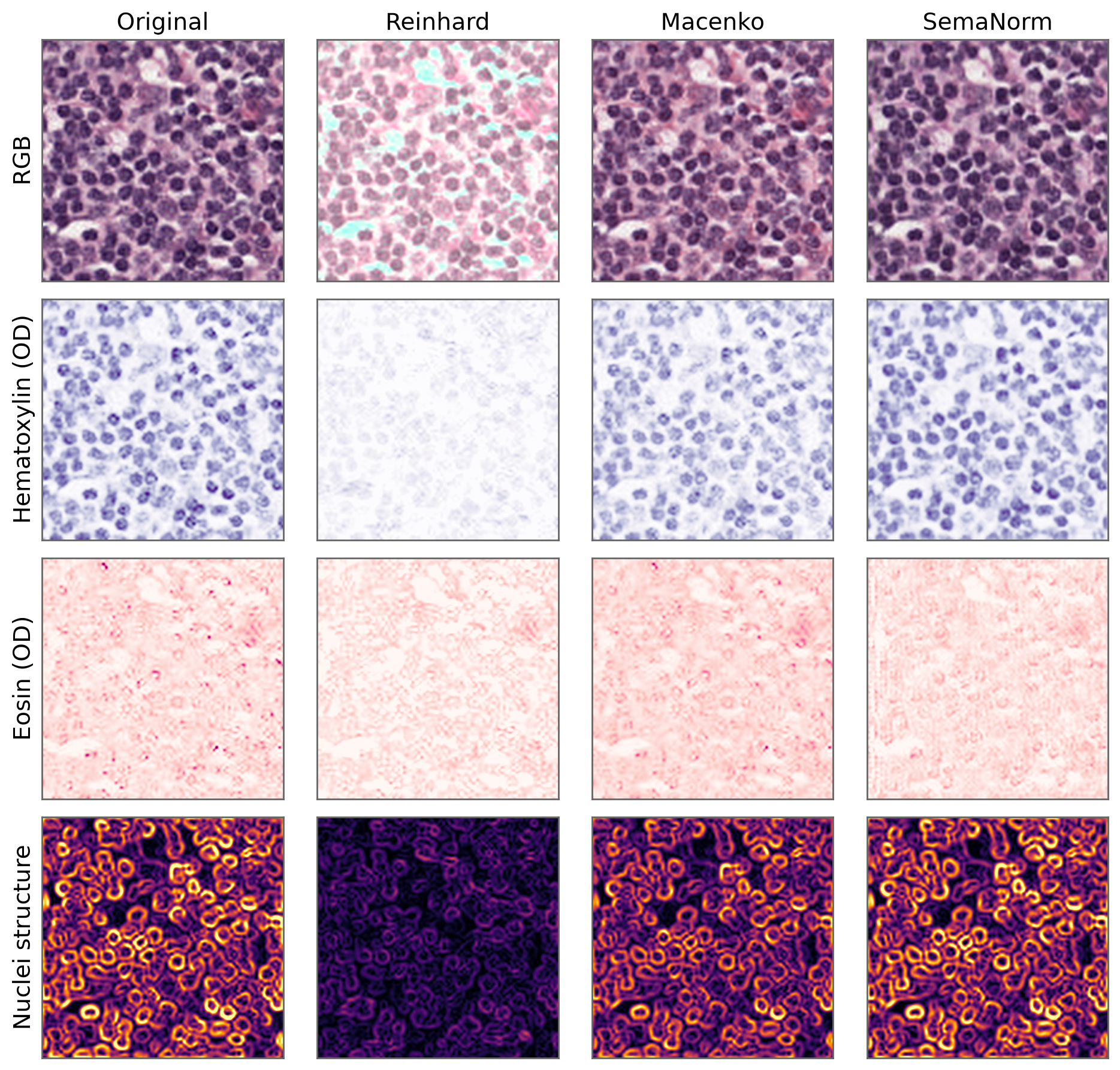}
\caption{Stain deconvolution \cite{Ruifrok2001ColorDeconvolution} and structure analysis for one patch. Rows show the RGB output, the hematoxylin and eosin optical-density channels, and the soft nuclei-structure map. SemaNorm reproduces the target stain balance while keeping the nuclei-structure map closest to the original, whereas the baselines shift the
hematoxylin/eosin balance and distort the structure map.}
\label{fig:decomp}
\vspace{-3mm}
\end{figure}

In Fig.~\ref{fig:qualitative} shows representative patches normalized by each method. SemaNorm reaches the canonical stain appearance while visibly preserving nuclei boundaries and glandular architecture, whereas the color-statistic baselines either under-correct the stain or flatten fine texture. In Fig.~\ref{fig:decomp} decomposes a single patch into its RGB output, hematoxylin and eosin optical-density channels, and soft nuclei-structure map: SemaNorm reproduces the target stain balance while keeping the nuclei-structure map closest to the original, consistent with the quantitative morphology results.

\section{Discussion and Limitations}
Even stain-robust foundation models exhibit measurable embedding shifts after conventional normalization (Table~\ref{tab:sem}), suggesting that these methods may alter semantic content beyond color appearance. SemaNorm reduces this shift while improving downstream performance for stain-sensitive models. Cell-graph evaluation is more reliable on the larger NCT-CRC patches, as the $96\times96$ Camelyon17-WILDS patches contain relatively sparse nuclei. Our semantic and morphology objectives also depend on the fidelity of the frozen encoders and nuclei segmenter, whose errors may propagate into the training signal. Finally, our evaluation covers identity, statistics-matching, stain-decomposition baselines, and two foundation models; future work will extend comparisons to learned normalizers such as StainGAN and StainNet and evaluate generalization across broader tissue acquisition domains.
\section{Conclusion}
We presented SemaNorm, a stain normalization method that makes morphology and
foundation-model semantics explicit training objectives, trained via stain-augmentation canonicalization to avoid both adversarial instability and
identity collapse. Across cross-hospital benchmarks SemaNorm delivered the
strongest out-of-distribution accuracy among the evaluated normalizers together with the best morphology preservation, and ablations confirmed the role of each component. The results argue that morphology-preserving normalization remains valuable even as foundation models become the dominant representation in computational pathology.
\bibliographystyle{IEEEtran}
\bibliography{references}

\end{document}